\documentclass{article}

\usepackage{microtype}
\usepackage{graphicx}
\usepackage{subfigure}
\usepackage{booktabs} % for professional tables
\usepackage{amsthm}
\theoremstyle{definition}
\newtheorem{definition}{Definition}[section]
\usepackage[T1]{fontenc}
\usepackage[utf8]{inputenc}
\usepackage{float}

\newcommand{\comment}[1]{}

\usepackage{hyperref}

\usepackage[accepted]{icml2021}

\icmltitlerunning{An Analysis of the Deployment of Models Trained on Private Tabular Synthetic Data: Unexpected Surprises}

\begin{document}

\twocolumn[
\icmltitle{An Analysis of the Deployment of Models Trained on Private Tabular Synthetic Data: Unexpected Surprises}

% It is OKAY to include author information, even for blind
% submissions: the style file will automatically remove it for you
% unless you've provided the [accepted] option to the icml2021
% package.

% List of affiliations: The first argument should be a (short)
% identifier you will use later to specify author affiliations
% Academic affiliations should list Department, University, City, Region, Country
% Industry affiliations should list Company, City, Region, Country

% You can specify symbols, otherwise they are numbered in order.
% Ideally, you should not use this facility. Affiliations will be numbered
% in order of appearance and this is the preferred way.
\icmlsetsymbol{equal}{*}

\begin{icmlauthorlist}
\icmlauthor{Mayana Pereira}{equal,to,goo}
\icmlauthor{Meghana Kshirsagar}{equal,to}
\icmlauthor{Sumit Mukherjee}{equal,to}
\icmlauthor{Rahul Dodhia}{to}
\icmlauthor{Juan Lavista Ferres}{to}
\end{icmlauthorlist}

\icmlaffiliation{to}{AI for Good Research Lab, Microsoft, Redmond, USA}
\icmlaffiliation{goo}{Department of Electrical Engineering, University of Brasilia, Brasilia, Brazil}

\icmlcorrespondingauthor{Mayana Pereira}{mayana.wanderley@microsoft.com}

% You may provide any keywords that you
% find helpful for describing your paper; these are used to populate
% the "keywords" metadata in the PDF but will not be shown in the document
\icmlkeywords{Differential Privacy,Synthetic Data,Machine Learning}

\vskip 0.3in
]

% this must go after the closing bracket ] following \twocolumn[ ...

% This command actually creates the footnote in the first column
% listing the affiliations and the copyright notice.
% The command takes one argument, which is text to display at the start of the footnote.
% The \icmlEqualContribution command is standard text for equal contribution.
% Remove it (just {}) if you do not need this facility.

%\printAffiliationsAndNotice{}  % leave blank if no need to mention equal contribution

\printAffiliationsAndNotice{\icmlEqualContribution} % otherwise use the standard text.

\begin{abstract}

Diferentially private (DP) synthetic datasets are a powerful approach for training machine learning models while respecting the privacy of individual data providers. The effect of DP on the fairness of the resulting trained models is not yet well understood. In this contribution, we systematically study the effects of differentially private synthetic data generation on classification. We analyze disparities in model utility and bias caused by the synthetic dataset, measured through %convention 
algorithmic fairness metrics.
Our first set of results show that although there seems to be a clear negative correlation between privacy and utility (the more private, the less accurate) across all data synthesizers we evaluated, more privacy does not necessarily imply more bias. Additionally, we assess the effects of utilizing synthetic datasets for model training and model evaluation. We show that results obtained % during model evaluation can deceive 
on synthetic data can misestimate the actual model performance when it is deployed on real data. We hence advocate on the need for defining proper testing protocols in scenarios where differentially private synthetic datasets are utilized for model training and evaluation.
 
  %Very recently, \cite{amazon, vitaly} have claimed that data generated via privacy preserving synthesizers might be more biased towards certain sub-groups. 
  %In this work we perform a systematic study of the effects of differentially private synthetic data generation in model bias introduced into the dataset, measured through convention algorithmic fairness metrics.
  %We find that, although there seems to be a clear negative correlation between privacy and utility (the more private, the less accurate) across all data synthesizers we evaluated, more privacy does not necessarily imply in more bias as some works have implied \cite{amazon}.
\end{abstract}

%\meghana{Don't need citations in abstract -- those are better in the intro. Abstract should be mainly about this work's contributions.}

%%
%% The code below is generated by the tool at http://dl.acm.org/ccs.cfm.
%% Please copy and paste the code instead of the example below.
%%

\section{Introduction}
%% This introduction is currently a copy and paste of other works!!!! please beware of that!! ideas here are to serve as a story line only!!

%% About privacy
Differential privacy (DP) is the standard for privacy-preserving statistical summaries \cite{dwork2006calibrating}. Companies such as Microsoft \cite{pereira2021us}, Google \cite{2020google}, Apple \cite{tang2017privacy}, and government organizations such as the US Census \cite{abowd2018us}, have successfully applied DP in machine learning and data sharing scenarios. The popularity of DP is due to its strong mathematical guarantees. Differential Privacy guarantees privacy by ensuring that the inclusion or exclusion of any particular individual does not significantly change the output distribution of an algorithm. 

%\meghana{Add more citations above}

%%About Synthetic Data
In areas ranging from health care, to education, and socioeconomic studies, the publication and sharing of data is crucial for scientific collaboration. However, the disclosure of such datasets can often reveal private, sensitive information. Privacy-preserving data publishing (PPDP) aims at enabling such collaborations while preserving the privacy of individual entries in the dataset. In the case of numerical/categorical data about individuals, the data can be modeled as a table, where each row represents information about an individual (say, details of their medical status or employment information). Privacy-preserving data publishing for such data can be done in the form of a synthetic data table that has the same schema and similar distributional properties as the real data. The aim here is to release a perturbed version of the original information, so that it can still be used for statistical analysis, but the privacy of individuals in the database is preserved.
The current state-of-the-art solution is to release differentially private datasets which come with DP guarantees for the individuals in the database. Informally, DP requires that what can be learned from the released data is approximately the same, whether or not any particular individual was included in the input database. This offers strong privacy protection and does not make any limiting assumptions about the power of the adversary: it remains a strong model even in the face of an adversary with additional background knowledge and reasoning power.

%% About Fairness
%In domains such as health care, education and consumer finance, the usage of machine learning models trained on sensitive data is ubiquitous \cite{chen2017disease,ghoddusi2019machine}. 
Machine learning models can have disparate impacts on minoritized subgroups \cite{wiens2019no,cohen2020predicting,rajotte2021reducing}. Unfairness in machine learning can happen, among other reasons, due to class imbalance and intrinsic bias in the underlying training dataset. It is known that differential privacy can affect fairness in machine learning models. However, despite significant work on addressing the relationship between differential privacy and ML fairness, there are fundamental questions that remain unanswered. For instance, it is not known how different synthetic data generation methods affect fairness. Also, in the literature, it is usually assumed that real data is available for testing models trained on synthetic data prior to deployment. This is not a realistic assumption. In many scenarios, only synthetic data is available during training and testing. So, it is important to study the performance of a model trained and tested with synthetic data vs its performance when tested against real data. In this paper we address these questions.

%These impacts often amplify existing prejudices against minorities in society under the guise of automated impartiality. In addition, as privacy laws get more rigorous around the world, machine learning models need to provide privacy guarantees to the individuals in the training data set. There are hence many open questions around the impacts of privacy-preserving techniques in model fairness.

\subsection{Related Works} Recent empirical studies have pointed out that differentially private deep learning has a disparate impact in subgroup accuracy \cite{vitaly}, and training models with differentially private synthetic images can increase subgroup disparities \cite{cheng2021can}. However these works have focused on image classification tasks where the disparity in accuracies are largely attributable to the class imbalance in these datasets: i.e disadvantaged classes are also rare classes in the dataset thereby leading to worse performance on these. In contrast, our work studies these issues in the context of tabular datasets and in settings where the data has an intrinsic bias against sub-populations that are not necessarily rare in the dataset.

\subsection{Contributions} In this work, we investigate the impacts of differentially private synthetic data on downstream classification, where we focus on understanding the impacts on model performance and fairness. % when utilizing differentially private synthetic data for model training. %We list our main contributions as follows:
We train models in two settings: (a) on real data and (b) on synthetic data and our evaluation finds that:

\begin{itemize}
    \item[1] When compared to models trained on real data, overall performance of models trained on synthetic data decreases as privacy increases. \textbf{However, the same does not hold for fairness metrics.} Additionally, we experiment with datasets that initially provide fair models, and observe that the synthetic data version results in unfair models.
    \item[2] \textbf{Models trained with differentially private synthetic data tend to perform \textit{more unfairly} when deployed on real data versus when tested on synthetic data.} 
\end{itemize}

To the best of our knowledge, our work is the first to evaluate the the fairness of machine learning models trained on DP synthetic data for the important case of tabular data. Moreover, it is also the first work that evaluates the potential disparities when utilizing synthetic data for training and testing models and later applying such models to real data.

\begin{figure*}[htp]
  \centering
  \label{img:method}
 \includegraphics[width=13cm]{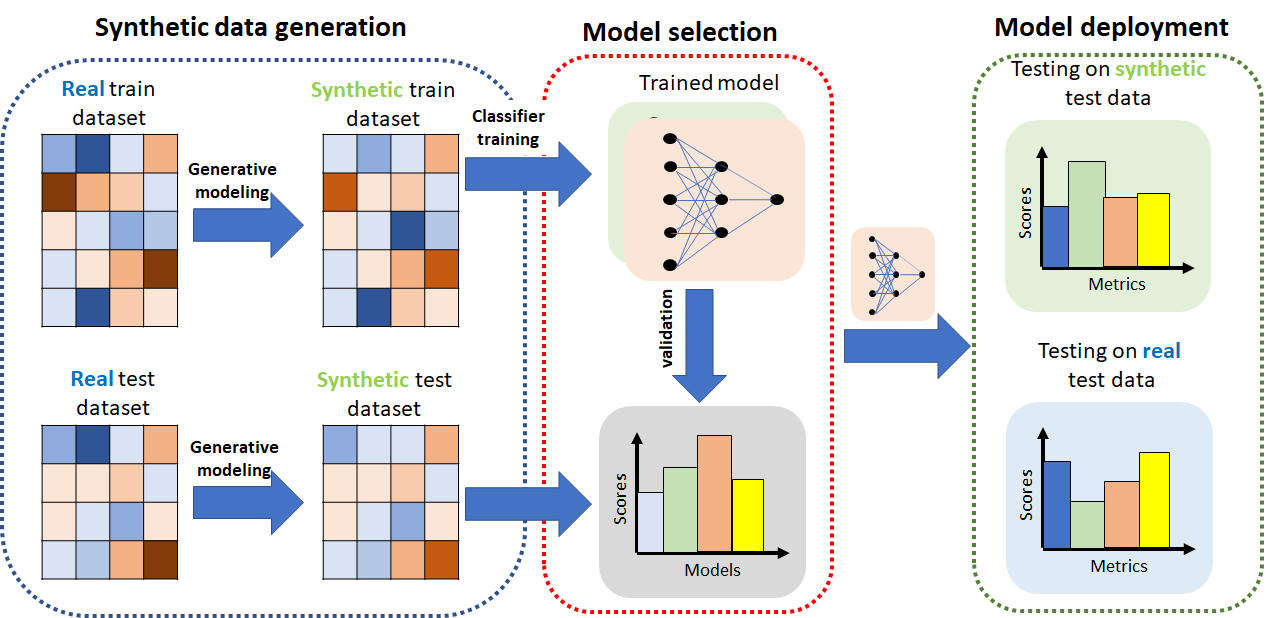}
  \caption{Pipeline for model training and evaluation using synthetic data  (1) We generate Synthetic datasets for model training and model testing utilizing differentially private synthesizers. (2) We train models utilizing synthetic data and evaluate on a synthetic test data. Model selection is made during this phase. (3) Based on the previous phase results, model is trained using synthetic data and deployed. Model is applied to real (test) data in production phase.}
\end{figure*}

\section{Preliminaries}

% Make modification in this sections. This section is currently heavily based on Zhu's paper
\subsection{Differential privacy}
Differential privacy is a rigorous privacy notion used to protect an individual’s data in a dataset disclosure. We present in this section notation and definitions %, and theorems 
that we will use to describe our privatization approach.  We refer the reader to \cite{book}, \cite{mcsherry} and \cite{calibrate} for detailed explanations of these definitions and theorems. % proofs.

\begin{definition}\textit{Pure Differential Privacy.} A randomized mechanism $\mathcal{M}:\mathcal{D}\rightarrow \mathcal{A}$ with data base domain $\mathcal{D}$ and output set $\mathcal{A}$ is $\epsilon$-differentially private if, for any output $A \subseteq \mathcal{Y}$ and neighboring databases $D, D' \in \mathcal{D}$ (i.e., $D$ and $D'$ differ in at most one entry), we have
\[
    \textnormal{Pr}[\mathcal{M}(D) \in A] \leq e^{\epsilon}\textnormal{Pr}[\mathcal{M}(D') \in A]
\]
\end{definition}

\begin{definition}\textit{Approximate Differential Privacy.} A randomized mechanism $\mathcal{M}:\mathcal{D}\rightarrow \mathcal{A}$ with data base domain $\mathcal{D}$ and output set $\mathcal{A}$ is $(\epsilon,\delta)$-differentially private if, for any output $A \subseteq \mathcal{Y}$ and neighboring databases $D, D' \in \mathcal{D}$ (i.e., $D$ and $D'$ differ in at most one entry), we have
\[
    \textnormal{Pr}[\mathcal{M}(D) \in A] \leq e^{\epsilon}\textnormal{Pr}[\mathcal{M}(D') \in A] + \delta
\]
\end{definition}

The privacy loss of the mechanism is defined by the parameter $\epsilon \geq 0$   in the case of 'pure' differential privacy and parameters $\epsilon, \delta \geq 0$ in the case of 'approximate' differential privacy.

The definition of neighboring databases used in this paper is user-level privacy. User-level privacy defines neighboring to be the addition or deletion of a single user in the data and all possible records of that user. Informally, the definition above states that the addition or removal of a single individual in the database does not provoke significant changes in the probability of any differentially private output. Therefore, differential privacy limits the amount of information that the output reveals about any individual.

A function $f$ (also called query) from a dataset $D \in \mathcal{D}$ to a result set $ A \subseteq \mathcal{A}$ can be made differentially private by injecting random noise to its output. The amount of noise depends on the sensitivity of the query. 

\subsection{Fairness Metrics}

In this section we present the definition of two different fairness metrics: Equal Opportunity \cite{heidari} and Statistical Disparity\cite{barocas}. Given a dataset $W = (X, Y', C)$ with binary protected attribute $C$ (e.g. race, sex, religion, etc), remaining decision variables $X$ and predicted outcome $Y'$, we define Equal Opportunity and Statistical Disparity as follows.
\comment{
\begin{definition}{Disparate Impact.} , we say that $C$ has disparate impact if 
\end{definition}
$$\frac{\textnormal{Pr}(Y'=1|C=0)}{\textnormal{Pr}(Y' =1|C=1)} \leq 0.8$$
}
 \begin{definition}{Equal Opportunity/ Equality of Odds} requires equal True Positive Rate (TPR) across subgroups:
\[
    \textnormal{Pr}(Y'=1|Y=1,C=0)=\textnormal{Pr}(Y'=1|Y=1,C=1)
\]
\end{definition}
where Y' is the model output.
\begin{definition}{Statistical Parity} requires positive predictions to be unaffected by the value of the protected attribute, regardless of true label 
\[
    \textnormal{Pr} (Y' = 1|C = 0) = \textnormal{Pr} (Y' = 1|C = 1)
\]
\end{definition}
We follow the approach of \cite{amazon, perrone2020fair} and utilize difference in Equal Oportunity (DEO) = $|\textnormal{Pr}(Y' =1|Y =1,C=0) - \textnormal{Pr}(Y' =1|Y =1,C=1)|$ and difference in Statistical Parity (DSP) = $|\textnormal{Pr} (Y' = 1|C = 0) - \textnormal{Pr} (Y' = 1|C = 1)|$ to measure model fairness. In our experiments with COMPAS dataset, we utilize True Negative Rate to quantify Equal Opportunity of a beneficial outcome.

%\meghana{Figure-1, Part-1, one of the blocks should contain Synthetic \textbf{Test} data?} -> Addressed this

\subsection{Differentially Private Synthetic Data Generators.}
We use several differentially private (DP) synthetic data generators that have been specifically tailored for generating tabular data with the goal of enhancing their utility for learning tasks. We consider two broad categories of approaches: i) Bayesian network based methods, ii) and Generative Adversarial Network (GAN) based models.

\subsubsection{Bayesian network based method}
\paragraph{PrivBayes}
In order to improve the utility of the generated synthetic data, \citep{zhang2017privbayes} try to approximate the actual distribution of the data by constructing a Bayesian network using the correlations between the data attributes. This allows them to factorize the joint distribution of the data into marginal distributions. Next, to ensure differential privacy, noise is injected into each of the marginal distributions and the simulated data is sampled from the approximate joint distribution constructed from these noisy marginals.

\subsubsection{GAN based methods}
\paragraph{QUAIL}
Quail-ified Architecture to Improve Learning (QUAIL) \cite{QUAIL} is an ensemble model approach that combines a DP supervised learning model with a DP synthetic data model to produce DP synthetic data. QUAIL framework can be used in conjunction with different synthesizers techniques. We introduce CTGAN, PATE methods, which  are the basis methods we utilize in our experiments with the ensemble approach of QUAIL. We note that unlike PrivBayes, both the QUAIL based approaches provide an approximate DP guarantee.

\textbf{Conditional Tabular GAN (CTGAN)} \cite{xu2019modeling} is an approach for generating tabular data. CTGAN adapts GANs by addressing issues that are unique to tabular data that conventional GANs cannot handle, such as the modeling of multivariate discrete  and mixed discrete and continuous distributions. It achieves these challenges by augmenting the training procedure with mode-specific normalization, by employing a conditional generator and training-by-sampling that allows it to explore discrete values more evenly. When applying differentially private SGD (DP-SGD) \cite{abadi2016deep} in combination with CTGAN the result is a DP approach for generating tabular data. This involves adding random noise to the discriminator and clipping the norm to make it differentially private.

The \textbf{PATE (Private Aggregation of Teacher Ensembles)} framework \cite{papernot2016semi} protects the privacy of sensitive data during training, by transferring knowledge from an ensemble of teacher models trained on partitions of the data to a student model. To achieve DP guarantees, only the student model is published while keeping the teachers private. The framework adds Laplacian noise to the aggregated answers from the teachers that are used to train the student models. CTGAN can provide  differential privacy by applying the PATE framework. We call this combination PATE-CTGAN, which is similar to PATE-GAN \cite{jordon2018pate}, for images. The original dataset is partitioned into $k$ subsets and a DP teacher discriminator is trained on each subset. Further, instead of using one generator to generate samples, $k$ conditional generators are used for each subset of the data.

\subsection{Datasets}
\paragraph{Adult Dataset} In the Adult dataset (32561 instances), the features were categorized as protected variable (C): gender (male, female); and response variable (Y): income (binary); decision variables (X): the remaining variables in the dataset. We map into categorical variables all continuous variables.
 
\paragraph{Prison Recidivism Dataset}
From the COMPAS dataset (7214 instances), we select severity of charge, number of prior crimes, and age category to be the decision variables (X). The outcome variable (Y) is a binary indicator of whether the individual recidivated (re-offended), and race is set to be the protected variable (C). We utilize a reduced set of features as proposed in \cite{calmon2017optimized}.
\paragraph{Fair Prison Recidivism Dataset} \label{fair_data}
We construct a "fair" dataset based on the COMPAS recidivism dataset by employing a data preprocessing technique for learning non-discriminating classifiers from \citep{kamiran2012data}, which involves changing the class labels in order to remove discrimination from the dataset. This approach selects examples close to the decision boundary to be either `promoted', i.e label flipped to the desirable class, or `demoted', i.e label flipped to the undesirable class (ex: the 'recidivate' label in the COMPAS dataset is the undesirable class). By flipping an equal number of positive and negative class examples, the class skew in the dataset is maintained.

\section{Experiments}
Our experiments follow a simple methodology in order to compare the fairness outcomes of logistic regression models when trained with real and synthetic data. We generate synthetic data using three differentially private synthesizers: PrivBayes \cite{privbayes}, QUAIL+PATECTGAN and QUAIL+DPCTGAN \cite{QUAIL}. For each synthetic data generation technique, we generate datasets utilizing different privacy budgets.

\subsection{Experimental Setup}
In our experiments we randomly divide the real dataset into an 80/20 split, separating the data into training and test datasets. We consider three synthesizers: i) PrivBayes, ii) QUAIL+PATECTGAN, and iii) QUAIL+DPCTGAN. For each synthesizer technique and for two values of privacy budget $\epsilon = 0.1, 10$, we run 10 rounds of synthetic DP data generation on the 80\% split. We utilize the SmartNoise Library\footnote{https://smartnoise.org} implementation of the QUAIL synthesizers, and approximate-DP approaches use the library's default value of $\delta$. For experiments using PrivBayes Synthesizers, we use the DiffPrivLib implementation \footnote{https://github.com/IBM/differential-privacy-library}. We train Logistic Regression and Random Forests models using the generated DP synthetic datasets. We test the trained models performance on the 20\% test split from the real data. For each model, we measure the following: area under the ROC curve (AUC); subgroup true positive rate (TPR); subgroup positive predictions. We report, for each technique and each value of privacy loss, the mean across 10 rounds. We apply the pipeline to the UCI Adult dataset \cite{adult} and ProPublica’s COMPAS recidivism data \cite{compass}, and the fair COMPAS dataset as defined in Section \ref{fair_data}. 

\paragraph{Evaluation of Model Performance on Synthetic Datasets vs. Real Datasets}

Differentially private synthetic datasets have the potential to be shared among researchers and the general public without compromising the privacy of the users in the dataset. As we envision a scenario where these datasets are used for model training (and also in the model evaluation phase) but the resulting models are applied to real data, as showcased in Figure \ref{img:method}, we ask ourselves what are the performance discrepancies between model evaluation phase (done with synthetic data) and model deployment phase (when model is deployed on real data). We investigate if there are in fact any clear discrepancies,  from the perspective of utility and fairness.

\subsection{Utility and Fairness Evaluation of Models Trained with Differentially Private Synthetic Datasets}\label{exp1}

\paragraph{Utility Evaluation} We evaluate the quality of the synthetic datasets for classification by measuring the area under the ROC curve (AUC). We compare the AUC obtained in our experiments with the AUC measured by training models with the real Adult and COMPAS datasets.

%%%%%%%%%%%%%%%%%%%%
%%%%%%%%%%%%%%%%%%%%
%%%% UTILITY DECREASE
%%%% auc variation

\begin{figure*}[ht]
\vskip 0.2in
\begin{center}
\includegraphics[width=.3\textwidth]{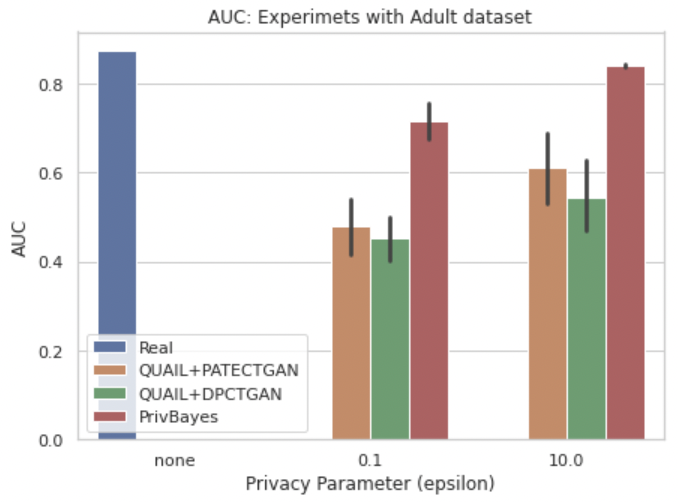}
\hfill
\includegraphics[width=.3\textwidth]{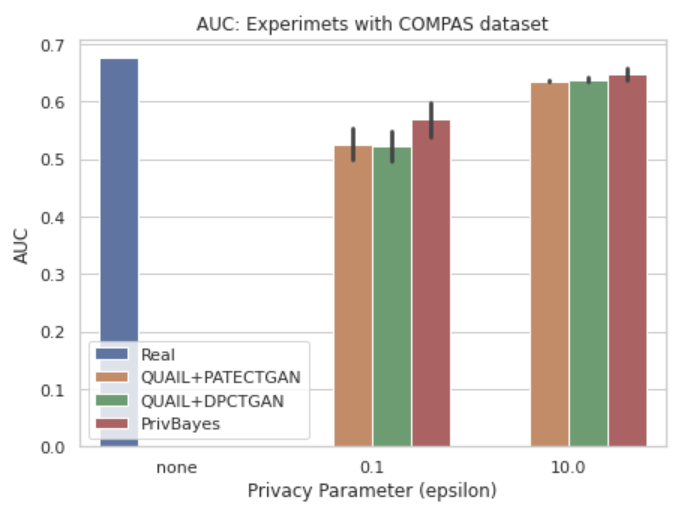}\hfill
\includegraphics[width=.3\textwidth]{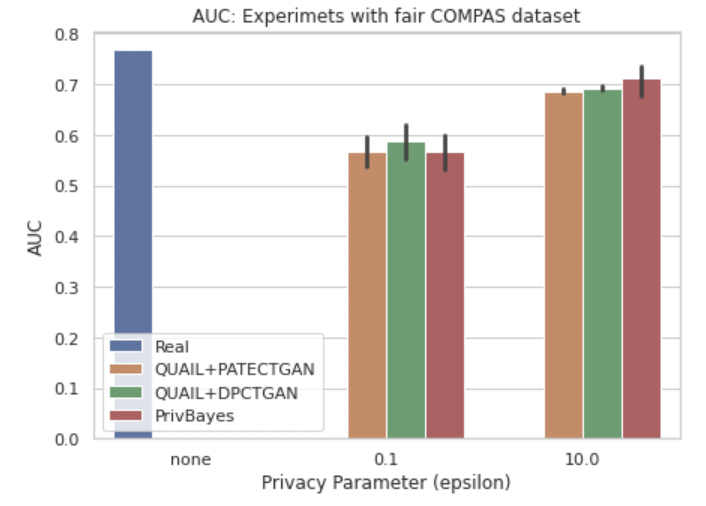}

\caption{Utility evaluation of classification models trained with differentially private synthetic datasets generated using PrivBayes and QUAIL techniques. Left shows AUC comparison of models trained using synthetic Adult datasets. Center shows the AUC comparison of models trained using synthetic COMPAS datasets. Right shows the AUC comparison of models trained using synthetic Fair COMPAS datasets.}
\label{fig:auc_exp1}
\end{center}
\vskip -0.2in
\end{figure*}

We confirm that lower epsilons yield models with lower performance. In the experiments with the Adult dataset ( Figure \ref{fig:auc_exp1}), models trained with data generated by the PrivBayes synthesizer perform significantly better than models trained with QUAIL generated synthetic datasets. If we increase the privacy guarantees (epsilon = 0.1), the models trained with QUAIL datasets become ineffective, with AUC $\approx 0.45$. An AUC lower than 0.5 indicates that the model performs worse than random guess. Experiments with COMPAS and Fair COMPAS datasets show a more balanced performance across models trained with data generated by the three synthesizers. At an $\epsilon = 10.0$, the PrivBayes synthesizer generated high-quality differentially private synthetic datasets for all three datasets.

%#######
%%%%%%%%%%%%%%%%%
%%%%%%%%%% FAIRNESS EVALUATION

\paragraph{Fairness Evaluation} We evaluate model bias and fairness by measuring the proportion of positive predictions in each class of the protected attribute, and the True Positive Rate (or True Negative Rate) in each class of the protected attribute. The Experiments with the Adult dataset show that synthetic data generated using PrivBayes can lead to slightly less biased and more fair models, as indicated in Table \ref{tab:adult_exp1}. Although QUAIL models also present less biased and True Positive Rates fairer, we would like to stress that for the experiments with the Adult dataset, the low AUCs indicates that models perform like random guesses, with little prediction value. 

\begin{table}[hbt!]
\caption{Bias and Fairness Metrics: Experiments with Adult, COMPAS and Fair COMPAS datasets. We evaluate the percentage of Positive Prediction in each class to evaluate model bias, and the True Positive Rate (TPR) or True Negative Rate (TNR) to evaluate model fairness. The classes in the two datasets are: Adult dataset - Female (F) and Male (M) and COMPAS dataset - African American (A) and Caucasian (C). These results are using Logistic Regression trained with real data (Real) and synthetic datasets generated with three synthesizers (Synth.): PrivBayes (PB), QUAIL+PATECTGAN (QP) and QUAIL+DPCTGAN (QC).}
\vskip 0.15in
\begin{center}
\begin{small}
\begin{sc}

  \label{tab:adult_exp1}
  \begin{tabular}{llcccc}
\toprule
 &&\multicolumn{2}{c}{Positive Pred.}&\multicolumn{2}{c}{TPR/TNR} \\
 \midrule
    Synth. & $\epsilon$  & F & M  & F & M\\
    \midrule
\multicolumn{6}{c}{Adult Dataset}\\
    \midrule
Real & --  & 0.06 & 0.22 & 0.41 & 0.53\\

PB & 10 & 0.06 & 0.2 & \textbf{0.43} & \textbf{0.49}\\
 &0.1 & 0.03 & 0.14 & 0.12 & 0.27\\
QP &10& 0.19 & 0.32  & 0.24 & 0.39\\
 &0.1&  0.24 & 0.23  & 0.24 & 0.22\\
QC & 10& 0.25 & 0.23  & 0.20 & 0.21\\
 &0.1&  0.6 & 0.56 & 0.57 & 0.52\\
\toprule
 Synth. & $\epsilon$  & A & C & A & C\\
    \midrule
\multicolumn{6}{c}{COMPAS Dataset}\\
    \midrule
   
Real & -- & 0.5 & 0.30 & 0.64 & 0.78\\
PB & 10&  0.61 & 0.40  & 0.50 & 0.67\\
 &0.1&  0.63 & 0.39 & 0.40 & 0.63\\
QP  &10&0.51 & 0.28  & 0.63 & 0.79\\
& 0.1& 0.44 & 0.47  & 0.59 & 0.56\\
QC  &10&  0.49 & 0.27 & 0.64 & 0.80\\
 &0.1&  0.46 & 0.43 & 0.55 & 0.59\\
\toprule
\multicolumn{6}{c}{Fair COMPAS Dataset}\\
 \midrule
Real & -- & 0.49 & 0.46 & 0.69 & 0.69\\
PB & 10  & 0.59 & 0.30 & 0.57 & 0.83\\%
 &0.1& 0.56 & 0.53 & 0.50 & 0.52\\%
QP  &10& 0.48 & 0.53 & 0.67 & 0.62\\%
  & 0.1& 0.45 & 0.47 & 0.62 & 0.58\\
QC  &10 & 0.46 & 0.57 & 0.71 & 0.57\\
 &0.1 & 0.47 & 0.49  & 0.62 & 0.57\\
  \bottomrule
\end{tabular}
\end{sc}
\end{small}
\end{center}
\vskip -0.1in
\end{table}

The evaluation of the models trained with synthetic COMPAS dataset show that models trained with PrivBayes synthetic datasets present results with similar bias (difference in Positive Predictions for the classes African-American and Causian is still similar), however, it increases the percentage of positive predictions of both groups in about 10 points percentage. This reflects in lower True Negative Rate numbers for both classes. For an $\epsilon = 10$, QUAIL models perform very similar to the real data in terms of bias and fairness. For $\epsilon = 0.1$, QUAIL models perform in a less-biased and fair manner, which we believe is mostly caused by the decrease in utility.

In the scenario of having a real dataset with less biased and more fair characteristics, we observe that QUAIL generates datasets that result in non-biased and fair classifier performance. Models trained with PrivBayes synthetic data, however, showed biased results. Interestingly, the metrics resemble the model trained with real COMPAS data. 

%\begin{figure}[H]
%\label{POS_compass_RF}
%\centering
%\includegraphics[width=8cm]{fairness_decrease.png}
%\caption{Positive prediction.}

%\end{figure}

The fairness observations with random forests models are very similar to the ones described above with logistic regression models. We also observe similar trends in Random Forest models and Logistic Regression models. One important point to make is that in scenarios with more privacy ($\epsilon = 0.1$), PrivBayes gets more unfair as models trained with QUAIL generated data tend to perform with more fairness.

\subsection{Utility and Fairness Disparities in Model Evaluation and Deployment}

We evaluate utility by measuring the differences in AUC measured during the evaluation phase, i.e. when model performance is evaluated using synthetic data, and the AUC measured during deployment phase , i.e. when the model is evaluated using real data. As in Section \ref{exp1}, we investigate the disparities between evaluation and deployment using three different datasets: Adult dataset, COMPAS dataset and Fair COMPAS dataset.

We find that models trained with PrivBayes synthetic datasets, in general, perform better on real data. AUC values are higher on real data, when compared to the same models evaluated on synthetic data. Table \ref{tab:auc_exp2} shows the evaluation AUC and deployment AUC comparisons. This result gives us an indication that when utilizing synthetic data generated using the PrivBayes techniques to train models, the expected model utility on deployed real data is higher than the model utility on synthetic test data.

However, our experiments indicate the same positive outcome is not the case for bias and fairness metrics. As we observe in Table \ref{tab:fairness_exp2} PrivBayes produced more biased models, and aside from the experiments with $\epsilon = 0.1$ to generate synthetic Adult datasets, all other experiments resulted in fairness deployment performance to be worse than evaluation performance.

\begin{table}[hbt!]
  \caption{Differences in evaluation AUC and deployment AUC. Positive differences refer to AUC during deployment higher than AUC during evaluation, and negative values mean that AUC during evaluation was higher.}
  \vskip 0.15in
\begin{center}
\begin{small}
\begin{sc}
  \label{tab:auc_exp2}
  \begin{tabular}{lcccc}

    \toprule
   
    &\multicolumn{2}{l}{Logistic Regression}&\multicolumn{2}{l}{Random Forests}\\
   \midrule
    &$\epsilon$ = 0.1&$\epsilon$ = 10 &$\epsilon$ = 0.1&$\epsilon$ = 10\\
 \midrule
     \multicolumn{5}{c}{Adult Dataset}\\
    \midrule
    PB & 0.19 & 0.05 & 0.17 & 0.04\\
    QP & -0.04 & 0.08 & -0.03 & 0.09\\
    QC & -0.12 & -0.04 & -0.07 & -0.04\\

     \midrule
     \multicolumn{5}{c}{COMPAS Dataset}\\
    \midrule
       PB & 0.02 & 0.08 & 0.02 & 0.06\\
    QP & 0.06 & -0.28 & 0.05 & -0.26\\
    QC &  - 0.47 & -0.32 & -0.48 &-0.36\\

  \bottomrule
\end{tabular}
\end{sc}
\end{small}
\end{center}
\vskip -0.1in
\end{table}

Models trained with QUAIL+CTGAN datasets, on the other hand, always present a decrease in utility from evaluation to deployment. The bias and fairness metrics also present a decrease in performance when comparing evaluation with deployment (except for experiments with adult dataset and $\epsilon = 0.1$. For this experiment, the utility of the model was so low that the model did not have any predictive power). 

%%%%%%%%%%%%%%%%%%%%%%
%%%%%%%%%%%%%%%%%%%%%%
%%%%%%%%%%%%%%%%%%%%%%

\begin{table}[hbt!]
  \caption{Fairness disparities in model evaluation and model deployment. Positive differences refer to AUC during deployment higher than AUC during evaluation, and negative values mean that AUC during evaluation was higher.}
  \vskip 0.15in
\begin{center}
\begin{small}
\begin{sc}
  \label{tab:fairness_exp2}
  \begin{tabular}{llcccc}
\toprule
 && \multicolumn{2}{c}{$\epsilon = 0.1$} & \multicolumn{2}{c}{$\epsilon = 10$} \\
 \midrule
    Data & S & PP & TPR & PP & TPR\\
    
        \midrule
& PB & $\downarrow$ & $\uparrow$ & $\downarrow$ & $\downarrow$\\
Adult & QP& $\uparrow$ & $\uparrow$ & $\downarrow$ & $\downarrow$\\
&QC& $\uparrow$ & $\uparrow$ & $\downarrow$ & $\downarrow$\\
    \midrule
& PB & $\downarrow$& $\downarrow$& $\downarrow$& $\downarrow$\\

COMPAS & QP & $\downarrow$& $\downarrow$& $\downarrow$& $\downarrow$\\

&QC  &$\downarrow$& $\downarrow$& $\downarrow$& $\downarrow$\\

  \bottomrule
\end{tabular}
\end{sc}
\end{small}
\end{center}
\vskip -0.1in
\end{table}

Finally, models trained with QUAIL+PATECTGAN did not present any specific pattern in terms of utility. As we can observe in Table \ref{tab:fairness_exp2}, in the Adult dataset experiments utility increases for high epsilons and and decreases for low epsilons. However, in the experiments with COMPAS and fair COMPAS the oposite pattern is observed. In terms of fairness QUAIL+PATECTGAN behaves in the same way as QUAIL+DPCTGAN.

\section{Limitations and Future Works}
The analysis we presented in this paper applies to synthetic data generated by the QUAIL and PrivBayes frameworks. It is interesting to see if similar results apply to other frameworks. The datasets used in our analysis are standard in the fairness literature. However, checking the validity of our results for larger, and high dimensional datasets is an interesting sequel to this work. 

%The real datasets in our experiments are not large in size ($\approx$30k samples Adult; $\approx$8k samples COMPAS) for big data standards and have low dimensionality. 
%We leave as future work a similar investigation with other synthesizer techniques and datasets with larger sample sizes and high dimensionality.
%Additionally, our experiments reveal the necessity of model testing protocols in scenarios where differentiallt private synthetic datasets are utilized for model training and evaluation.

\section{Conclusion}
As the privacy-preserving research community develops new and more sophisticated techniques for privacy-preserving data publishing, the natural question of fairness impacts arises. This paper investigates the implications in model fairness when utilizing differentially private synthetic data for model training. We observe that model utility continuously decreases as we increase the privacy guarantees of synthetic data. However, fairness performance seems to be synthesizer dependent. Additionally, we observe that models trained with differently private synthetic data tend to perform more unfairly when tested on real data versus when tested on synthetic data. This is an important observation as we see synthetic data techniques becoming more accepted as the standard data publishing approach in domains such as health care, education, and other population studies.

\bibliography{example-paper}
\bibliographystyle{icml2021}

\end{document}